\documentclass[10pt,twocolumn,letterpaper]{article}

\usepackage[pagenumbers]{cvpr} 

\usepackage[dvipsnames]{xcolor}

\definecolor{cvprblue}{rgb}{0.21,0.49,0.74}
\usepackage[pagebackref,breaklinks,colorlinks,citecolor=cvprblue]{hyperref}
\usepackage{graphicx}
\usepackage{algorithm}
\usepackage{amssymb}
\usepackage{amsmath}
\usepackage{mathrsfs}
\usepackage{algorithmicx}
\usepackage{algpseudocode}
\usepackage{textcomp}
\usepackage{multirow}
\usepackage{booktabs}    

\def\paperID{15565} 
\def\confName{CVPR}
\def\confYear{2024}

\title{Segment Anything in Defect Detection}

\author{Bozhen Hu\textsuperscript{a,b}, Bin Gao\textsuperscript{c}, 
Cheng Tan\textsuperscript{b}, Tongle Wu\textsuperscript{d}, and Stan Z. Li\textsuperscript{b}\thanks{Corresponding author}\\
\tt\small $^a$ Zhejiang University, Zhejiang, China;
\tt\small $^b$ Westlake University, Zhejiang, China;\\
\tt\small $^c$ University of Electronic Science and Technology of China, Chengdu, China; \\
\tt\small $^d$ Pennsylvania State University, School of Electrical Engineering and Computer Science \\
\tt\small hubozhen@westlake.edu.cn, \tt\small bin$\_$gao@uestc.edu.cn, \\
\tt\small tancheng@westlake.edu.cn, \tt\small tfw5381@psu.edu,
\tt\small stan.zq.li@westlake.edu.cn
}
\begin{document}
\maketitle
\begin{abstract}
Defect detection plays a crucial role in infrared non-destructive testing systems, offering non-contact, safe, and efficient inspection capabilities. However, challenges such as low resolution, high noise, and uneven heating in infrared thermal images hinder comprehensive and accurate defect detection. In this study, we propose DefectSAM, a novel approach for segmenting defects on highly noisy thermal images based on the widely adopted model, Segment Anything (SAM)~\cite{kirillov2023segany}. Harnessing the power of a meticulously curated dataset generated through labor-intensive lab experiments and valuable prompts from experienced experts, DefectSAM surpasses existing state-of-the-art segmentation algorithms and achieves significant improvements in defect detection rates. Notably, DefectSAM excels in detecting weaker and smaller defects on complex and irregular surfaces, reducing the occurrence of missed detections and providing more accurate defect size estimations. Experimental studies conducted on various materials have validated the effectiveness of our solutions in defect detection, which hold significant potential to expedite the evolution of defect detection tools, enabling enhanced inspection capabilities and accuracy in defect identification.
\end{abstract}

\section{Introduction}
\label{sec:intro}
Image segmentation is a fundamental task in computer vision that involves dividing an image into meaningful and distinct regions or objects~\cite{wu2023image,ghosh2019understanding}. It plays a crucial role in various applications, including object recognition, scene understanding, medical imaging, autonomous driving~\cite{wang2022medical}, etc. The goal of image segmentation is to assign a label or category to each pixel in the image, grouping similar pixels together based on their visual characteristics. Image segmentation algorithms have evolved over the years, ranging from traditional techniques like thresholding and region growing~\cite{mancas2005segmentation} to more advanced deep learning-based approaches using convolutional neural networks (CNNs) and semantic segmentation models~\cite{matterport_maskrcnn_2017, jjfaster2rcnn, DBLP-abs-1802-02611}. These advancements have significantly improved the accuracy and efficiency of image segmentation, opening up new possibilities for visual perception and intelligent systems.

\begin{figure*}[htbp!]
\centering
\includegraphics[width=0.97\textwidth]{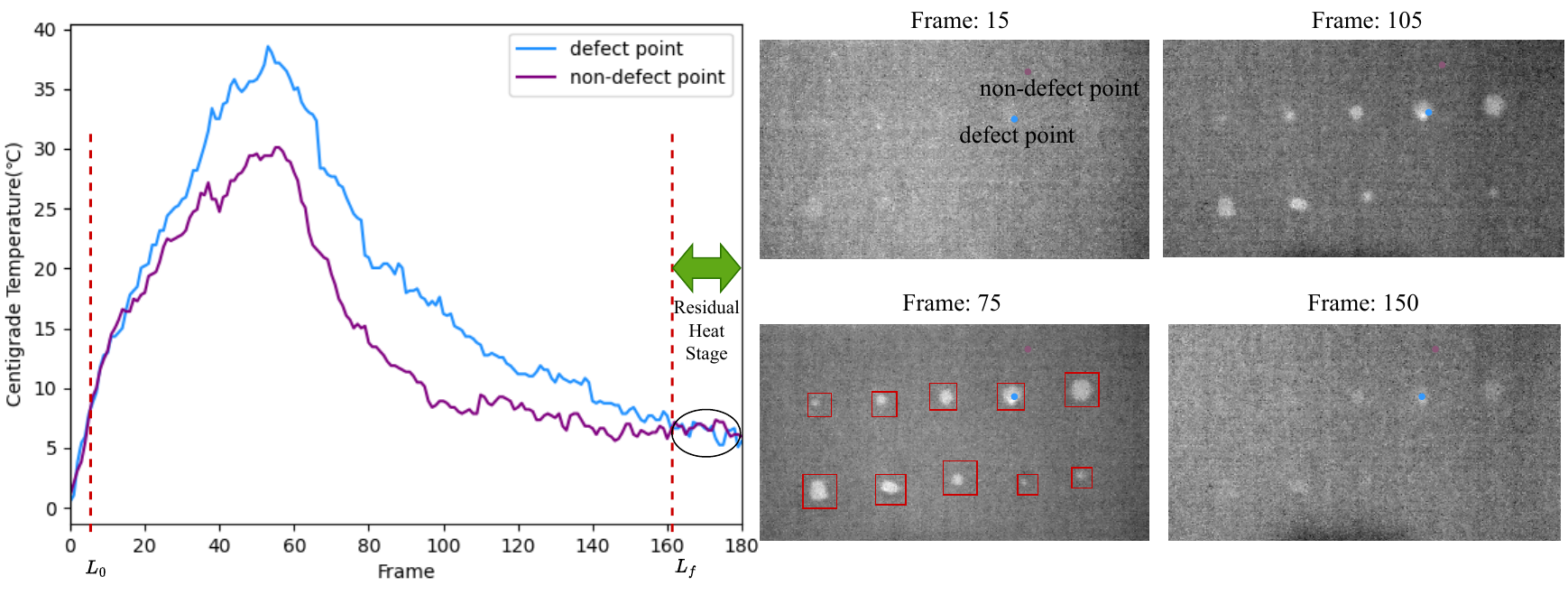} 
\caption{Temperature change curve at the defect and non-defect points, along with corresponding infrared thermal images captured by the OPT system. Defects are highlighted in red rectangles for frame 75. The temperature disparity between the defect and non-defect areas enables effective discrimination. $L_0$, warming-up frame; $L_f$, cooling-off stage; Residual Heat Stage is defined in section~\ref{Dataset}. }
\label{fig-curve1}
\end{figure*}

Nondestructive testing (NDT)~\cite{cartz1995nondestructive} refers to a range of analysis techniques employed in the fields of science and technology to assess the characteristics of materials, components, or systems without inducing any harm or damage. Optical pulsed thermography testing (OPT) is a crucial technology in NDT, offering advantages such as fast speed, non-contact operation, and safety~\cite{fernandes2016carbon, ciampa2018recent}. OPT systems have found extensive applications in detecting defects in carbon fiber reinforced polymers across various industries, including aerospace, shipbuilding, automotive, civil engineering, and sports equipment~\cite{kudo2023identification}, etc. In laboratory experiments, we utilize OPT and portable optical pulsed thermography (POPT) systems to conduct thermal experiments on different carbon fiber-reinforced polymer materials, generating and collecting infrared thermal data over the years. The fundamental principle of defect detection using infrared thermal imaging involves heating a specimen with external light, which induces the transmission of heat energy within the specimen. Variations in power between defective and non-defective areas indicate the presence of defects. The surface temperature of the specimen changes over time, providing valuable information for determining the locations of defects in the infrared thermal images, which exhibit distinct heat values. Fig.~\ref{fig-curve1} illustrates the temperature change curve and corresponding thermal images, demonstrating that defects are easily distinguishable when the temperature difference is significant (e.g., Frame: 75), while they become more challenging to identify at the initial (e.g., Frame: 15) and final stages (e.g., Frame: 150). In this study, our objective is to accurately detect all defects, even in high noise thermal images with small temperature gaps, and estimate precise defect areas. 

Traditional approaches, such as using the discrete Fourier Transform~\cite{ibarra2005interactive} or principal component analysis~\cite{rajic2002principal}, have been employed to enhance the contrast between targets and background by reducing or eliminating noise. However, deep learning models have shown significant promise in image segmentation due to their ability to learn intricate image features and deliver precise segmentation results~\cite{hu2020lightweight, Yakubovskiy:2019}. Nevertheless, supervised and semi-supervised models often suffer from overfitting issues when applied to thermal images, which exhibit similarities across the overall dataset and cannot be effectively addressed in practice. To bridge the gap between thermal image segmentation and universal applications, we draw inspiration from recent advancements in natural image segmentation, which have introduced segmentation foundation models~\cite{kirillov2023segany, ma2023segment}, known for their exceptional adaptability and performance across diverse segmentation tasks. Leveraging the pre-training and fine-tuning paradigm, as well as prompts-guided designs, we have developed DefectSAM, a specific thermal image segmentation model that works in complex conditions, such as uneven heating on irregular surfaces. The approach incorporates expert participation to effectively reduce missed defect detection and improve defect size estimation. In summary, our contributions include:
\begin{itemize}
    \item Constructing extensive defect thermal databases from two optical pulsed thermography systems, encompassing various materials and releasing them.
    \item Developing DefectSAM, the first large-scale foundation model for defect inspection, surpassing state-of-the-art (SOTA) segmentation foundation models and demonstrating superior performance even in complex scenarios.
    \item Characterizing the criteria for defect detection, identifying urgent challenges in this field, and providing solutions, suggestions, and directions.
\end{itemize}

\section{Related Work}
\label{sec:Related}
In this section, we present related works to introduce image segmentation and defect detection.
\subsection{Image Segmentation}
In the field of image segmentation, several related works have contributed to advancing state-of-the-art techniques. One prominent approach is the Fully Convolutional Network (FCN)~\cite{Shelhamer2014FullyCN}, which laid the foundation for subsequent developments in semantic segmentation. Instance segmentation has also seen significant progress with the introduction of Mask R-CNN~\cite{He2017MaskR} by extending the framework of Faster R-CNN~\cite{Ren2015FasterRT}. Another notable advancement is the U-Net architecture~\cite{Ronneberger2015UNetCN}, which has been widely adopted and has shown exceptional performance in medical image segmentation tasks~\cite{Zhou2018UNetAN, isensee2021nnu, shen2017deep}. Additionally, there have been efforts to incorporate attention mechanisms~\cite{wang2018non} to capture long-range dependencies and improve segmentation performance, and transformers are found better under end-to-end finetuning~\cite{goldblum2023battle}. Different from previous multi-task segmentation systems that combine semantic, instance, and panoptic segmentation tasks together~\cite{Zhang2021KNetTU,Cheng2021MaskedattentionMT,Jain2022OneFormerOT}, Segment Anything (SAM)~\cite{kirillov2023segany} is a versatile model that performs promptable segmentation tasks, allowing it to adapt to a wide range of existing and new segmentation tasks through prompt engineering. This capability enables SAM to generalize its performance across various tasks, although not all tasks can be accommodated. Prompt engineering is also fit to tackle the problems in defect detection to find weak and small defects and generalize to new thermal images. 

\subsection{Infrared Thermal Defect Detection}
Typical unsupervised algorithms, such as PCA~\cite{tong2022quantitative}, PPT~\cite{qiao2021ppt}, TSR~\cite{moskovchenko2020detecting}, and matrix decomposition methods~\cite{DBLP:conf/icip/Wu0W21}, are  commonly employed to enhance the signal-to-noise ratio and reduce data dimensions. These techniques aim to minimize noise and extract pertinent information from the data, enabling manual inspection to identify defect locations. Furthermore, deep learning algorithms~\cite{medak2022defectdet} have been utilized for defect detection and area calculation. Many of these algorithms have been enhanced by mainstream object detection and semantic segmentation methods~\cite{xue2023memory}. Given the visual resemblance between infrared and medical images, defects can be likened to tumors in the human body (see Fig.~\ref{fig-defect-tumor}), leading to the application of medical image segmentation methods in this domain~\cite{hu2020lightweight, ma2023segment, Zhou2018UNetAN}.

\begin{figure}[htbp!]
  \centering
   \includegraphics[width=0.9\columnwidth]{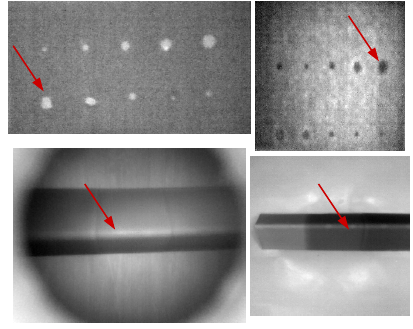}
   \caption{Defects in thermal images obtained from both the OPT and POPT (bottom left) systems are depicted, showcasing various types of materials. In each thermal image, the location of a single defect is indicated by a red arrow.}
   \label{fig-defect_types}
\end{figure}

However, the limitations of these algorithms arise due to the similarities in appearance among thermal images, the lack of diverse datasets and materials, and the time-consuming and labor-intensive nature of laboratory experiments. Consequently, deep learning-based models often suffer from heavy overfitting and limited generalization capabilities, restricting their applicability to unseen types of defects. Fig.~\ref{fig-defect_types} shows defects in different thermal images.

\section{Method}
\label{3_Method}
This section interprets the problem definitions, data curation and pre-processing methods, network and training protocols, and metrics on thermal images.
\subsection{Task}
In the field of defect detection, there are four pressing challenges that need to be addressed urgently. (a) It is crucial to develop effective methods to identify all defects in thermal images, even in the presence of high levels of noise. (b) Enhancing the generalization ability of trained models is essential to ensure their applicability in detecting defects across different specimens. (c) Accurately determining the real sizes of defects using thermal images remains a significant concern. (d) Calculating the depth of defects in various materials requires further investigation. In this study, our focus is on addressing problems (a) and (b) through image segmentation techniques while also estimating more accurate defect segmentation areas for the problem (c). The challenge of calculating defect depth (problem (d)) will be explored in future research endeavors. 

\subsection{Dataset Creation and Pre-processing}
\label{Dataset}
Laboratory experiments are conducted using a high-precision OPT system and a flexible POPT system, as described in Appendix~\ref{platform}. The specimens used in the experiments are detailed in Appendix~\ref{specimen}, and they are categorized into two groups: flat-type samples and irregular type (R-type) samples. Infrared thermal imaging experiments are performed to generate three-dimensional volumetric data denoted as $D_i\in\mathbb{R} ^{m_i\times n_i\times f_i }$, where $(m_i, n_i)$ represents the spatial dimensions of the $i$-th data, and $f_i$ is the total number of frames, $i\in [1,\cdots,B]$. By adjusting experimental settings such as sampling positions, classes, and excitation time, we can obtain diverse volumetric thermal data, which is essential for addressing problems (a) and (b). 

For two-dimensional image segmentation, we set a sampling interval $L_I$ to extract thermal images from each volumetric data, where $L_I=1$ indicates that all frames are used. Additionally, we define warming-up and cooling-off frames, denoted as $L_0$ and $L_f$ respectively. During the initial and final stages, the temperature approaches the surrounding ambient temperature, resulting in minimal temperature differences between defect and non-defect points (as shown in Fig.~\ref{fig-curve1}). Therefore, the initial $L_0$ frames and final $L_f$ frames are excluded from the analysis. Consequently, the total number ($N_f$) of thermal images in the constructed dataset is calculated as:
\begin{equation}
N_f(f_i)=B\frac{f_i-L_0-L_f}{L_I} 
\end{equation}

\begin{figure}[htbp!]
  \centering
   \includegraphics[width=0.9\columnwidth]{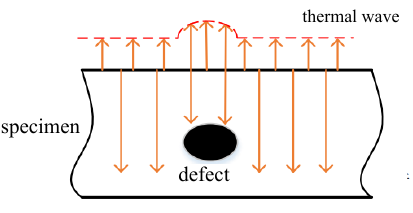}
   \caption{Schematic diagram of heat wave propagation in a material containing a defect.}
   \label{fig-wave}
\end{figure}

During pulsed thermographic inspection (refer to Fig.~\ref{fig-system}), heat takes place from the surface to the interior of the sample, leading to a gradual decrease in temperature. The thermophysical properties of the defect area play a significant role in influencing the heat conduction process, resulting in a temperature disparity between the defective region and the surrounding background areas, as illustrated in Fig.~\ref{fig-wave}.
\begin{figure*}[htbp!]
\centering
\includegraphics[width=0.8\textwidth]{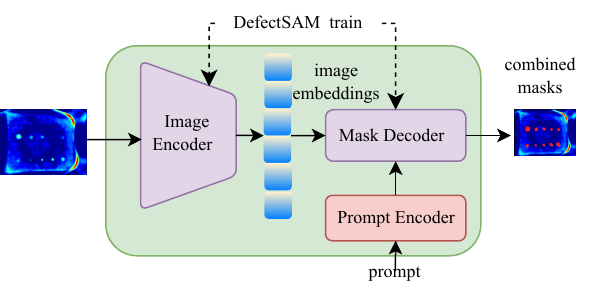} 
\caption{Pipeline of the DefectSAM. }
\label{fig-model}
\end{figure*}

Temperature variation at $z$ from the inner surface of the specimen can be calculated as follows:
\begin{equation}
T(z, t)=\frac{Q}{\sqrt{\pi \rho c k t}} \exp \left(-\frac{z^2}{4 \alpha t}\right)
\end{equation}
where $Q$ represents the input energy, $T$ denotes the temperature, $\rho, c, k$ refer to the density, heat capacity, and thermal conductivity of the material, $\alpha$ is the thermal diffusivity, and $t$ is the time~\cite{WANG2020106196}. $T(0, t)$ represents the surface temperature of the specimen. If there exists a defect inside the specimen with a depth of $d$, the corresponding surface temperature is denoted as $T_d(0, t)$. The temperature gap $\Delta T$ between defective and non-defective regions can be expressed as:
\begin{equation}
\label{eq(3)}
\Delta T=T_d(0, t)-T(0, t)=\frac{2 Q}{\sqrt{\pi \rho c k t}} \exp \left(-\frac{d^2}{\alpha t}\right)
\end{equation}
Eq.~\ref{eq(3)} is derived under standard experimental conditions. Combining the theoretical analysis mentioned above with deep learning techniques holds promise for addressing the problem (d) in the future. 

\noindent
\textbf{Definition 1 (Residual Heat Stage)} 
It refers to the period towards the end of a laboratory test, characterized by the last $L_f$ frames in the volumetric data $D_i$, where temperatures $T$ in both the defect and non-defect areas are nearly equal. However, these temperatures remain higher than the initial ambient temperature, as shown in Fig.~\ref{fig-curve1}. 

Based on Definition 1, we pre-process data $D_i$ to reduce the influence of residual heat and background noise. 
\begin{equation}
    x_{ij}=D_{ij}- \frac{{\textstyle \sum_{k=f_i-L_f+1}^{f_i}}D_{ik}}{L_f}
\end{equation}
where $x_j$ is the $j$-th frame of data $D_i$, and $j\in \left [ L_0+1,f_i-L_f\right ] $.
\subsection{Pre-taining and Fine-tuning in Defect Detection}
Problem (a) and problem (b) pose significant challenges to the advancement of deep learning models in defect detection within the field. To address these issues, a consensus has been reached on the need for deep learning model-assisted approaches, combined with expert involvement in the decision-making process regarding defect location. The concept of prompt learning~\cite{zhou2022learning, ding2021openprompt} has emerged as a driving force behind the implementation and progress of this methodology. Due to the inherent characteristics of defect images, such as high noise levels, similarity in patterns, and limited sample availability, conventional deep learning models often suffer from overfitting and lack generalization capabilities, which also appear in other fields. To mitigate these concerns, a pre-train and fine-tune paradigm for large-scale models has been established in the fields of computer vision and natural language processing. This approach involves initially training a model on a source task to acquire relevant knowledge, which is then fine-tuned on a new target task with limited labels, thereby enhancing the model's learning capacity. This mechanism is suitable to tackle the generalization problem. 

\begin{figure}[htbp!]
  \centering
   \includegraphics[width=0.8\columnwidth]{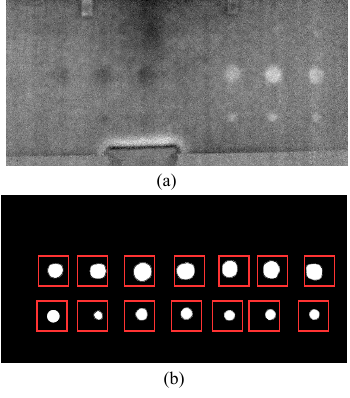}
   \caption{A thermal image and its labels. (a) The thermal image. (b) The semantic ground truth with prompts.}
   \label{fig-image-label}
\end{figure}

Notably, SAM~\cite{kirillov2023segany}, which has built the most extensive segmentation dataset to date, exhibits promptability and enables zero-shot transfer to new image distributions and tasks. Building upon the SAM model, we propose the DefectSAM pipeline, as depicted in Fig.~\ref{fig-model}, which comprises an image encoder, a prompt encoder, and a mask decoder. The vision transformer~\cite{dosovitskiy2020image} (ViT)-based image encoder plays a crucial role in extracting inherent features from the input images, enabling the model to understand and analyze the visual information effectively. Drawing inspiration from the success of powerful existing methods~\cite{he2022masked, radford2021learning}, the prompt encoder encompasses various prompts, such as text, boxes, points, and masks, facilitating human decision-making and interaction. The mask decoder, a lightweight module, consists of only two transformer layers~\cite{Vaswani2017AttentionIA} that fuse the image embedding and prompt encoding, enabling the model to generate precise and detailed defect masks. Pre-training is conducted through SAM, leveraging the scale of the data, the model itself, and the design of the prompt encoder to meet the requirements of an exceptional defect detection model. Fine-tuning of the image encoder and mask encoder is performed using the aforementioned defect dataset. As the prompts used for thermal images resemble those employed for universal images, we retain their original weights, which provide sufficient indications. An example of a thermal image and its ground truth with prompts (boxes) is shown in Fig.~\ref{fig-image-label}.

In the context of defect detection, as depicted in Fig.~\ref{fig-image-label}, our primary objective is to accurately segment all defective areas and precisely locate the defects within a thermal image. It is important to note that thermal images for defect detection typically consist of only two categories: background noise and defects. This distinction sets the specific requirements for defect detection apart from other models like SAM~\cite{kirillov2023segany} and MedSAM~\cite{ma2023segment}, which are designed to handle images containing various objects. 

\subsection{Training Protocol and Evaluation}
\begin{figure}[htbp!]
  \centering
   \includegraphics[width=0.9\columnwidth]{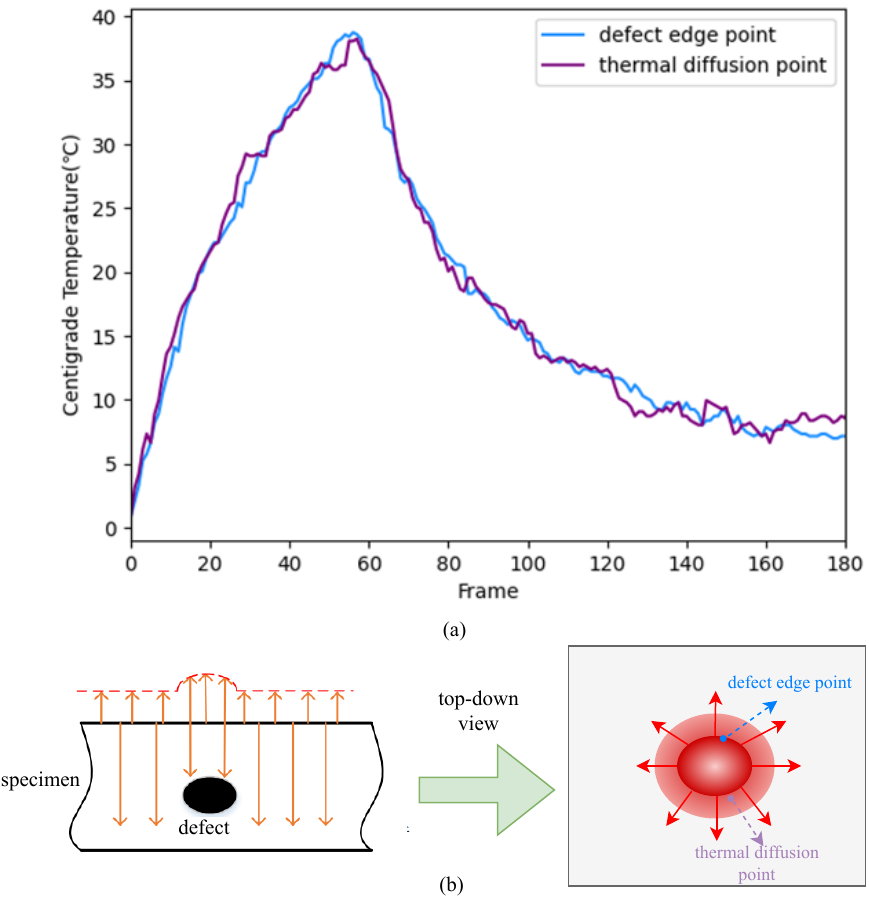}
   \caption{The thermal diffusion effect. (a) Temperature change curves of the defect edge point and the thermal diffusion point. (b) Thermal diffusion effect. Blue point: the defect edge point; purple point: background point influenced by the thermal diffusion.}
   \label{fig-diffusion}
\end{figure}

Regarding the labeling method, as detailed in Appendix~\ref{specimen}, materials containing defects are predominantly created by human beings, while some defects naturally occur due to the fatigue effect~\cite{GUO20224773}. The experimenters were aware of the approximate defect locations within the materials beforehand. To ensure accurate annotations, three experienced human annotators were engaged to independently annotate the original thermal image sequences or the images processed by certain unsupervised algorithms~\cite{tong2022quantitative}. Their annotations were then consolidated through a voting process to arrive at the final decisions. However, it is important to note that determining the actual size of defects present in thermal images is extremely challenging due to the influence of thermal diffusion. In detail, the observed signals are not the expected feature signals alone, as they are affected by interference caused by lateral and longitudinal thermal diffusion~\cite{levec1985longitudinal}. As illustrated in Fig.~\ref{fig-diffusion}, the temperature curve of a background point influenced by thermal diffusion closely resembles that of a genuine defect point. Consequently, for the problem (c), we tackle this issue by segmenting and calculating the defective areas in thermal images, providing an estimation of the real size of defects despite the presence of inaccurate labels resulting from thermal diffusion.

The loss function $L$ used in this context combines the Dice loss $L_\mathrm{Dice}$ and the binary cross-entropy loss $L_\mathrm{BCE}$, with weights assigned to each component. This combination has been demonstrated to be effective and reliable across various segmentation tasks~\cite{goldblum2023battle}. This hybrid loss is defined as:
\begin{equation}
\begin{aligned}
L & = L_{\operatorname{BCE}}+\beta L_{\operatorname{Dice}} \\
L_{\operatorname{BCE}} & =-\frac{1}{N} \sum_{n=1}^N y_{n} \log \hat{y}_{n} \\
L_{\operatorname{Dice}} & =1- \frac{2\sum_{n=1}^Ny_n\hat{y}_n}{\sum_{n=1}^N(y_n)^2+\sum_{n=1}^N(\hat{y}_n)^2}
\end{aligned}
\end{equation}  
where $y_n\hat{y}_n$ represent the ground truth of and predicted label of voxel $n$, $N$ is the total number of voxels in a thermal image, $\beta$ is a hyper-parameter, adjusting the weight of binary
cross-entropy and Dice losses.

For the problem (c), as mentioned earlier, obtaining the actual sizes of defects from thermal images is extremely challenging due to the influence of thermal diffusion. Therefore, we assess the performance of algorithms by computing the areas of defects compared to the ground truth. The Intersection Over Union (IOU) is commonly used as an evaluation metric in most semantic segmentation tasks. The IOU is calculated using the following formula:
\begin{equation}
\text{IOU} =\frac{\sum_{n=1}^N y_n  \hat{y}_n }{\sum_{n=1}^N (y_n)^2 + \sum_{n=1}^N (\hat{y}_n)^2-\sum_{n=1}^N y_n  \hat{y}_n} 
\end{equation}  
In the field of defect detection, the F-score is frequently employed to evaluate Precision ($P$) and Recall ($R$). It is defined as:
\begin{equation}
F = \left(\gamma^2+1\right)(P \times R) /\left(\left(\gamma^2 \times P\right)+R\right) 
\end{equation}
Precision calculates the ratio of true positive predictions (correctly predicted positive instances) to the total number of positive predictions (both true positives and false positives), focusing on how many of the predicted positive instances are actually correct. Recall calculates the ratio of true positive predictions to the total number of actual positive instances (both true positives and false negatives), emphasizing how many of the actual positive instances are successfully captured by the model.
In the above equation, $\gamma$ represents a parameter that balances the importance of precision and recall. In the thermal imaging debond diagnostic, a value of $\gamma$ is chosen as 2, indicating that Recall is prioritized over Precision. This decision is motivated by the objective of detecting all defects to prevent potential harm to materials and equipment. Neglecting even a single defect could lead to significant losses. In practical scenarios, it may be acceptable to compromise on accuracy to ensure comprehensive defect detection.

\section{Experiments}
\label{Experiments}
In this section, we present the experimental results, analysis, and observations.
\subsection{Dataset and Settings}
For each volumetric data, $D_i\in\mathbb{R} ^{m_i\times n_i\times f_i }$, the original height and width $(m_i, n_i)$ of $D_i$ are (480, 640) outputted from OPT system and (384, 288) from POPT system. The spatial size of input images is resized to (1024, 1024). In general, Typically, $f_i$ is greater than 180. We set $L_0=15, L_f=15$, and $L_I=5$. There are a total of $B=218$ volumetric data from these different types of samples. Consequently, the defect dataset contains over 6000 instances, but the segmentation ground truths are the same for each $D_i$. To create a diverse training set, we randomly split the dataset into 80$\%$ for training, 10$\%$ for validation, and 10$\%$ for testing. To assess DefectSAM's performance on various samples, we employ five-fold cross-validation to calculate the IOU and F-score.

The network is trained on the base SAM model and is optimized using the AdamW optimizer~\cite{Loshchilov2017FixingWD} with a default initial learning rate of $10^{-4}$ and a weight decay of 0.01. The batch size is set to 4, and $\beta$ is assigned a value of 1. The training process is performed on a single NVIDIA-SMI A100 GPU using PyTorch 1.13+cu117 with CUDA 11.2. During validation, the best checkpoint is selected as the final model. 

\subsection{Baselines}
To assess the performance of DefectSAM, we conduct a comparative analysis with several well-known image segmentation deep learning models. These models include H-DenseUNet~\cite{Li2017HDenseUNetHD}, TernusNet~\cite{Iglovikov2018TernausNetV2FC}, RVOS~\cite{Ventura2019RVOSER}, UNet++~\cite{Zhou2018UNetAN}, and an improved lightweight three-dimensional image segmentation model based on UNet++~\cite{hu2020lightweight}, H-DenseUNet and RVOS also predict defects in the spatiotemporal domain. Besides, few-shot learning image segmentation methods are considered, like ASGNet~\cite{Li2021AdaptivePL}, and~\cite{xue2023memory}. Additionally, we compare DefectSAM with the base model SAM~\cite{kirillov2023segany} and MedSAM~\cite{ma2023segment}, the latter is based on SAM and specifically designed for universal medical image segmentation. Given the three-dimensional nature of the defect data, the models designed for volumetric data segmentation utilize the same test set as the image segmentation methods. The key distinction lies in the sampling interval $L_I$. For video segmentation models, the sampling interval is set to $L_I=1$, whereas for image segmentation models, it remains the at $L_I=5$.

\subsection{Results}
\begin{table*}[tp]
  \caption{Visualization results of thermal images with prompts and predicted segmentation outputs.}
  \centering
  \begin{tabular}{lccccc}
    \toprule
    No. & Input & 3D-UNet++~\cite{hu2020lightweight} & SAM~\cite{kirillov2023segany} & MedSAM~\cite{ma2023segment} & DefectSAM \\
    \midrule
    (1) & \includegraphics[width=0.15\textwidth]{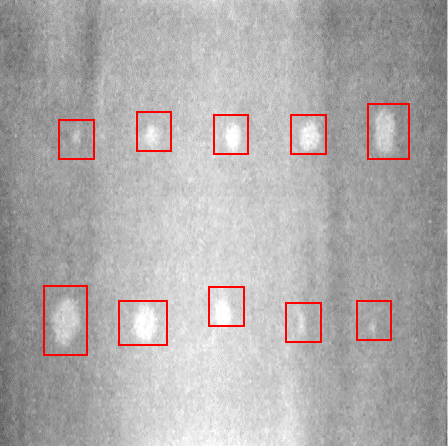} & \includegraphics[width=0.15\textwidth]{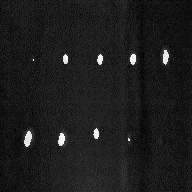} & \includegraphics[width=0.15\textwidth]{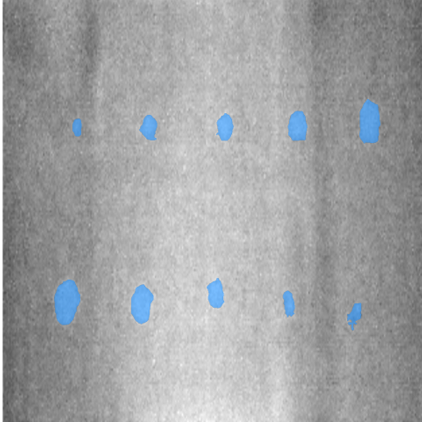} & \includegraphics[width=0.15\textwidth]{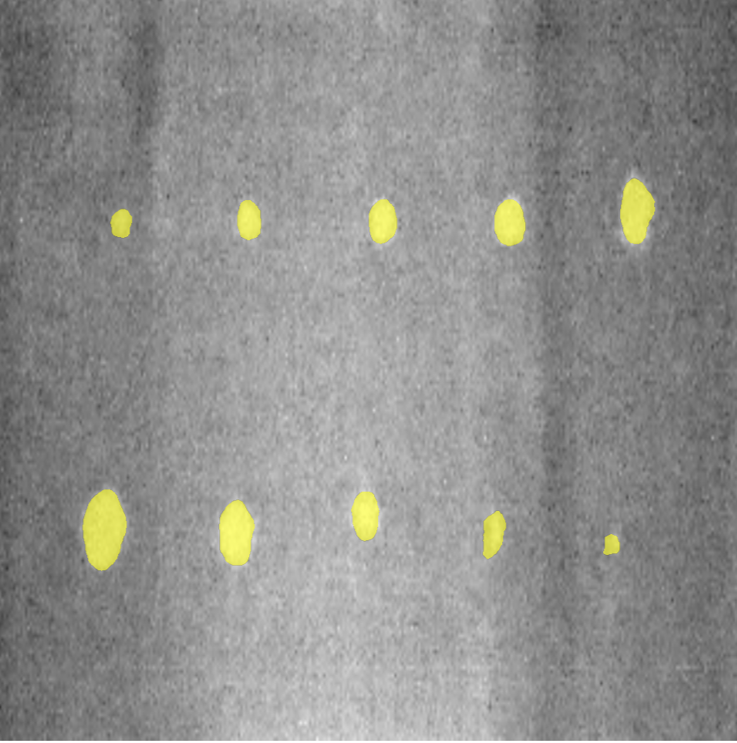} & \includegraphics[width=0.15\textwidth]{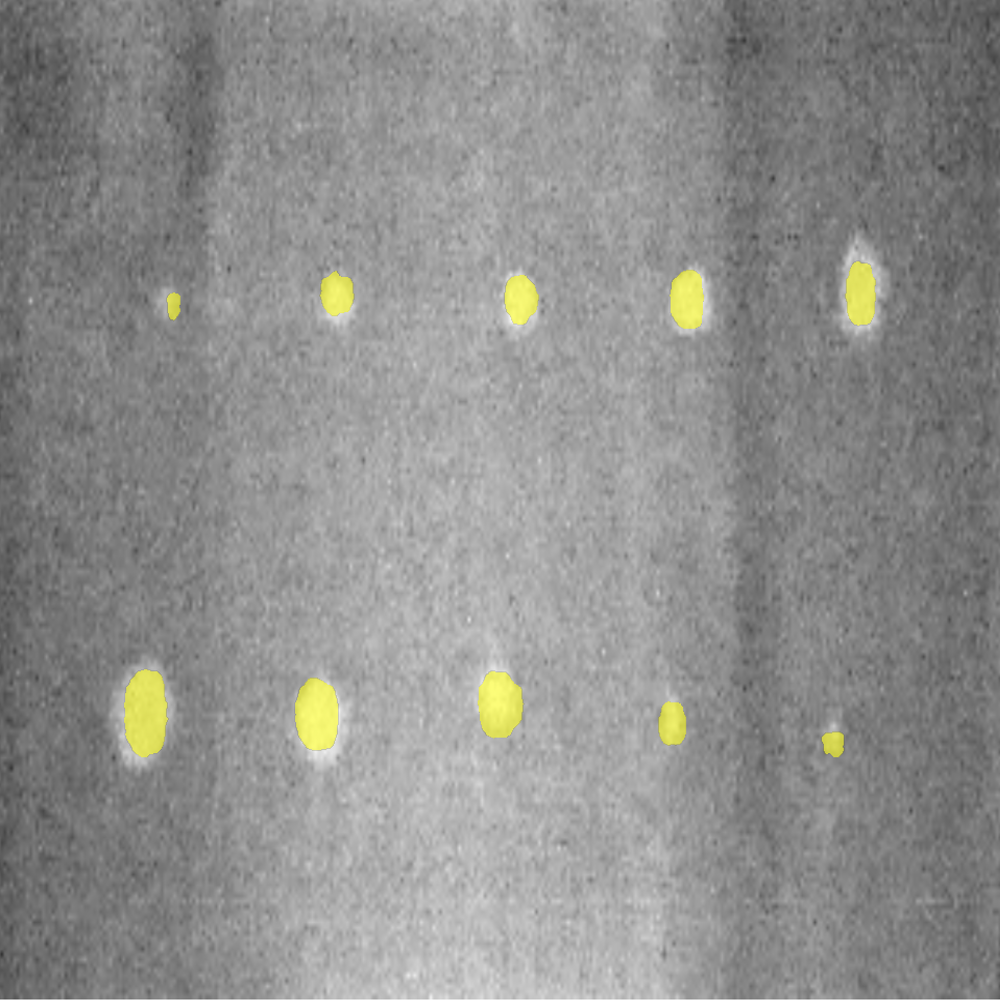} \\ 
    (2)& \includegraphics[width=0.15\textwidth]{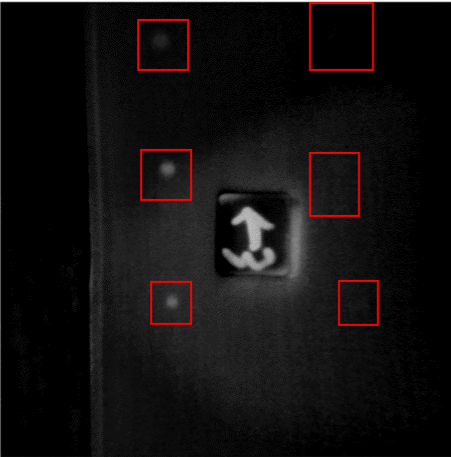} & \includegraphics[width=0.15\textwidth]{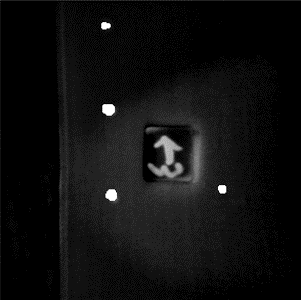} & \includegraphics[width=0.15\textwidth]{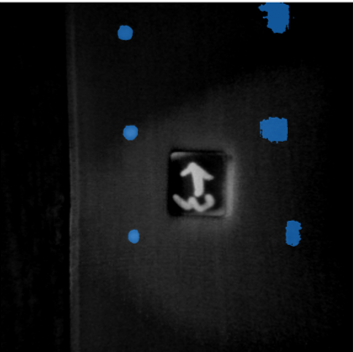} & \includegraphics[width=0.15\textwidth]{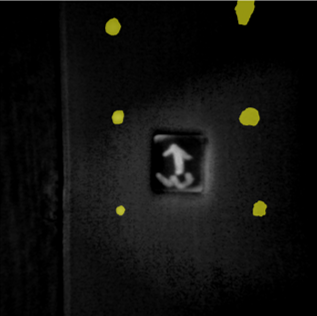} & \includegraphics[width=0.15\textwidth]{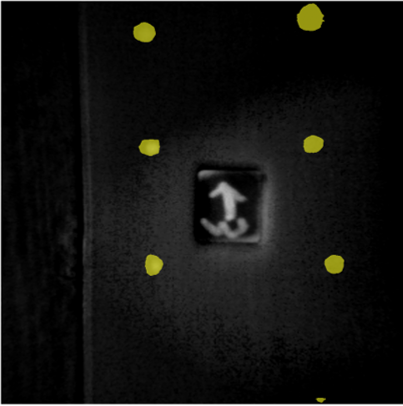} \\
    (3)& \includegraphics[width=0.15\textwidth]{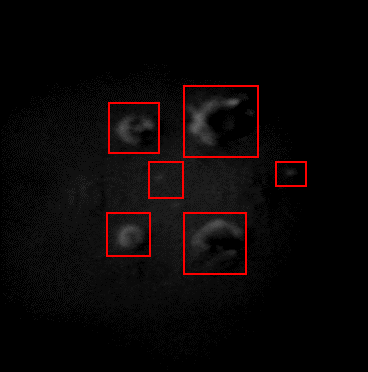} & \includegraphics[width=0.15\textwidth]{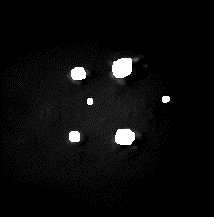} & \includegraphics[width=0.15\textwidth]{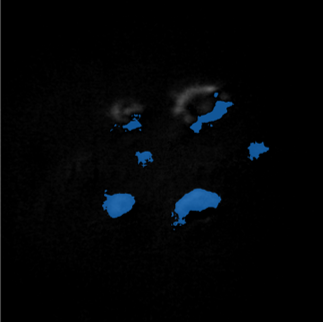} & \includegraphics[width=0.15\textwidth]{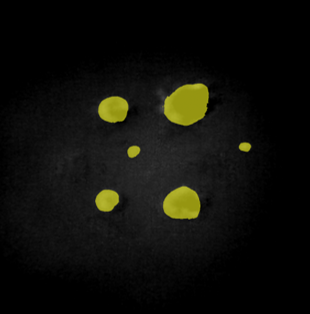} & \includegraphics[width=0.15\textwidth]{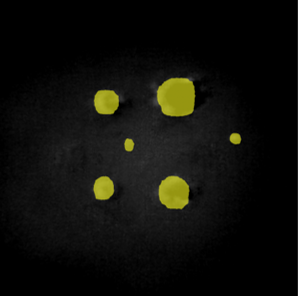} 
    \\
    (4)& \includegraphics[width=0.15\textwidth]{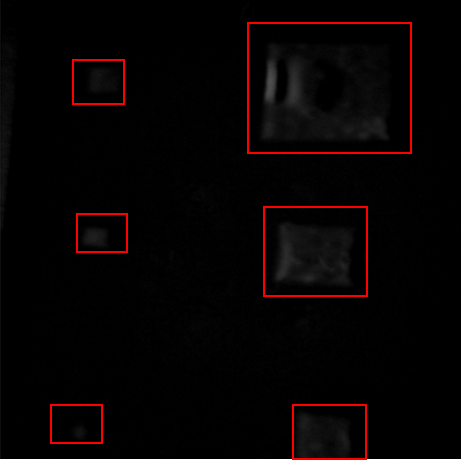} & \includegraphics[width=0.15\textwidth]{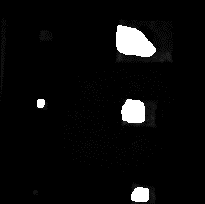} & \includegraphics[width=0.15\textwidth]{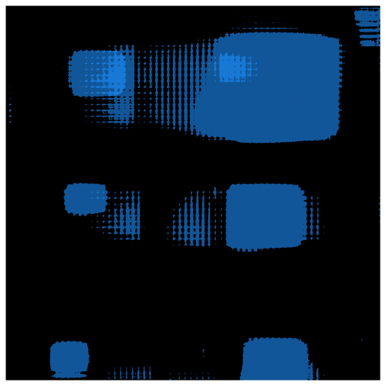} & \includegraphics[width=0.15\textwidth]{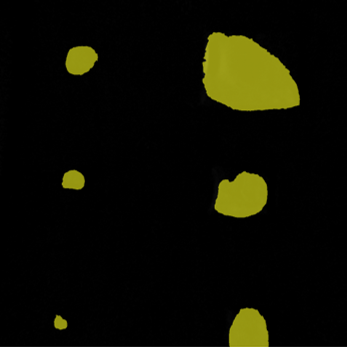} & \includegraphics[width=0.15\textwidth]{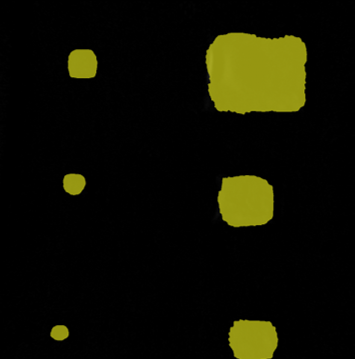} \\
    (5)& \includegraphics[width=0.15\textwidth]{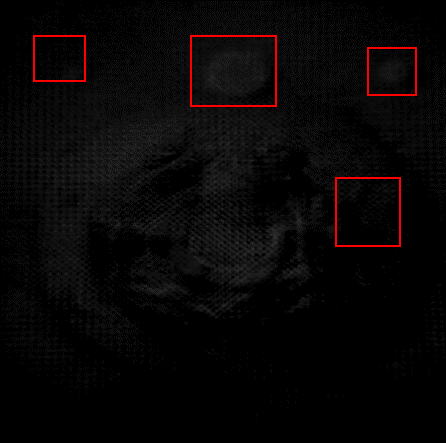} & \includegraphics[width=0.15\textwidth]{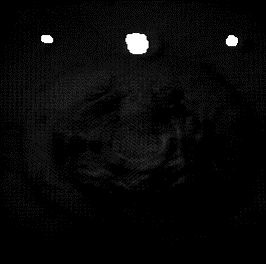} & \includegraphics[width=0.15\textwidth]{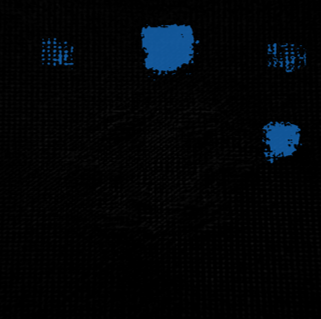} & \includegraphics[width=0.15\textwidth]{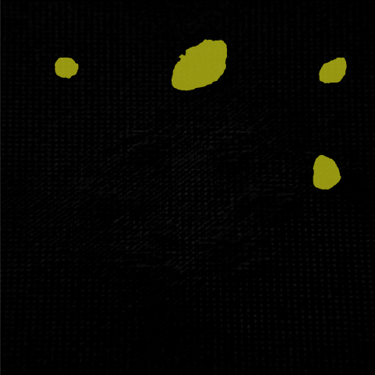} & \includegraphics[width=0.15\textwidth]{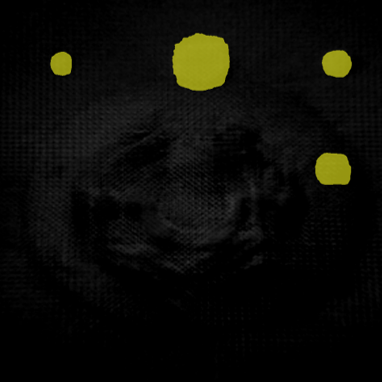} \\
    (6)& \includegraphics[width=0.15\textwidth]{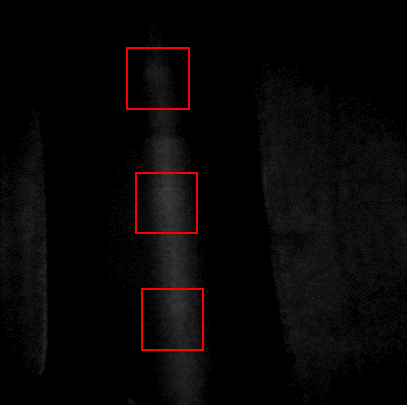} & \includegraphics[width=0.15\textwidth]{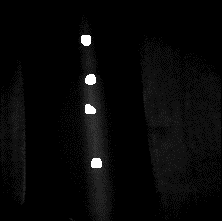} & \includegraphics[width=0.15\textwidth]{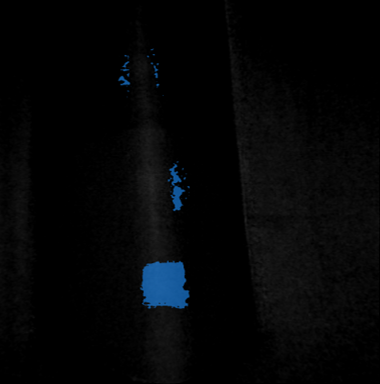} & \includegraphics[width=0.15\textwidth]{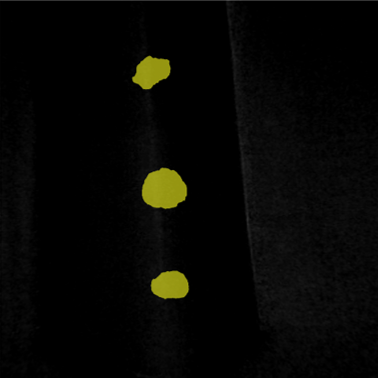} & \includegraphics[width=0.15\textwidth]{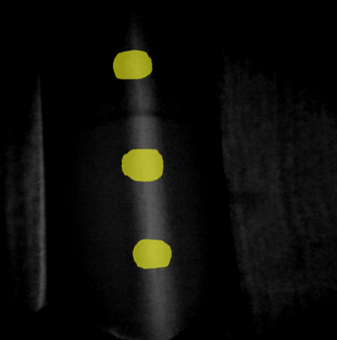} \\
    \bottomrule
  \end{tabular}
\label{table-visualization}
\end{table*}
In order to gain a deeper understanding of image segmentation in defect detection, we employed a visualization technique to showcase the thermal images alongside their predicted results, with segment labels overlaid on the original images. The dataset we constructed was divided into two distinct groups: flat-type data and R-type data. The comparison results are presented in Table~\ref{table-visualization}, where we list and number the input images, which are posted along with their corresponding prompts. The first five rows of images belong to the flat-type group, while the last row represents the irregular R-type. Notably, the number of R-type samples is smaller compared to the flat-type samples. Data for No.(1) was obtained from the OPT system, while the rest were collected from the POPT system. It is evident that the input for No.(1) contains more pronounced defective information, making it easily distinguishable. Additionally, the defect in the No.(3) image takes on a rectangular shape, whereas the others exhibit circular shapes.

Analyzing Table~\ref{table-visualization}, we observe that 3D-UNet++~\cite{hu2020lightweight} exhibits poor performance on these unseen thermal images. Defects in images No.(1), No.(2), No.(4), and No.(5) are not fully detected, indicating limited generalization ability. On the other hand, the universal image segmentation model, SAM~\cite{kirillov2023segany}, also misses some defects in the No.(6) image but performs relatively better. In terms of MedSAM~\cite{ma2023segment} and our proposed DefectSAM, both models successfully detect all defects in these thermal images, including small defects with high levels of noise. However, when comparing the two, DefectSAM produces more natural and accurate segmented shapes. For instance, MedSAM tends to segment all defects into circles, as seen in the No.(3) image, despite their actual shapes being rectangular. In contrast, our proposed model, DefectSAM, achieves successful defect detection while generating better-segmented shapes, highlighting its ability to address problems (a) and (b). To further illustrate these findings, we have calculated and included the evaluated metrics.

\begin{table}[h]
  \caption{F-score ($\%$) comparisons on the thermal image dataset. $\left [ ^* \right ]$ and $\left [ ^\dagger \right ]$ mean the results are taken from~\cite{hu2020lightweight} and \cite{xue2023memory}, respectively. The best results are shown in bold.}
  \centering
  \begin{tabular}{lccc}
    \toprule
    Method& Flat-type & R-type \\
    \midrule
    TernusNet~\cite{Iglovikov2018TernausNetV2FC}$^*$ & 81.94 & 11.78  \\
    H-DenseUNet~\cite{Li2017HDenseUNetHD}$^*$ & 83.97 & 9.73 \\
    RVOS~\cite{Ventura2019RVOSER}$^*$ & 76.10 & 48.38  \\
    UNet++~\cite{Zhou2018UNetAN}$^*$ & 91.54 & 82.44  \\
    3D-UNet++~\cite{hu2020lightweight}$^*$ & 93.33 & 88.70 \\ 
    ASGNet~\cite{Li2021AdaptivePL}$^\dagger$ & 68.46 & 63.36 \\
    \cite{xue2023memory}$^\dagger$ & 88.93 & 80.95 \\
    \hline
    SAM~\cite{kirillov2023segany} & 98.24 & 68.02 \\
    MedSAM~\cite{ma2023segment} & \textbf{100.0} & \textbf{100.0} \\ 
    DefectSAM (Ours) & \textbf{100.0} & \textbf{100.0}  \\
    \bottomrule
  \end{tabular}
\label{table-F}
\end{table}

\begin{figure}[htbp!]
  \centering
   \includegraphics[width=0.9\columnwidth]{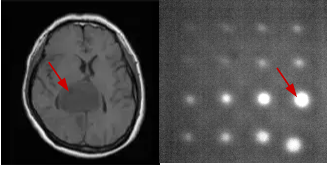}
   \caption{Magnetic Resonance Imaging (MRI) images~\cite{RANJBARZADEH2023106405} and thermal images, with the red arrows pointing at the locations of the tumor and one defect.}
   \label{fig-defect-tumor}
\end{figure}

In Table~\ref{table-F}, we present a comparison of F-scores for two types of thermal images obtained from the OPT and POPT systems. The R-type specimens, which are irregular in nature, pose a greater challenge in defect determination compared to the flat-type specimens. Consequently, in Table~\ref{table-F}, most of the methods exhibit lower performance on the R-type data. Notably, SAM~\cite{kirillov2023segany} achieves significantly better results on the flat-type data and comparable results on the R-type group, showcasing its strong image segmentation capabilities across diverse domains. As previously mentioned, the medical images exhibit semantic similarities with the collected defective images (refer to Fig.~\ref{fig-defect-tumor}). Therefore, both the medically enhanced SAM model, MedSAM~\cite{ma2023segment}, and our proposed model, DefectSAM, successfully and comprehensively detect all defects. These findings demonstrate that, with the aid of expert prompts, DefectSAM holds promise for real-life applications by accurately identifying defects and possessing favorable generalization ability. Consequently, it addresses the long-standing challenges (a) and (b) that have hindered the widespread adoption of deep learning models in the defect detection field.

In addressing the problem (c), we focus on segmenting the defective areas and utilize the Intersection over Union (IOU) as the evaluation metric. Table~\ref{table-IOU} presents the results of various comparison methods, including 3D-UNet++~\cite{hu2020lightweight}, its Sequence PCA improved version, and ASGNet~\cite{Li2021AdaptivePL}, \cite{xue2023memory}, all of which were trained using a thermal defect dataset. Due to the presence of thermal diffusion, the defect information is susceptible to being overshadowed by noise. Consequently, these methods yield relatively lower IOU values when assessed using pixel-level image segmentation evaluation metrics. It is important to note that SAM~\cite{kirillov2023segany} and MedSAM~\cite{ma2023segment} have not been exposed to the defect dataset prior to evaluation. However, they achieve comparable results with the aforementioned models trained on thermal images or volumetric data. Remarkably, MedSAM achieves the same F-score values as DefectSAM. However, upon examining Table~\ref{table-visualization}, we observe that MedSAM's segmented shapes of defects exhibit certain issues. This phenomenon becomes more apparent when considering Table~\ref{table-IOU}, where MedSAM's IOU values are significantly lower than those of the proposed DefectSAM. In comparison, DefectSAM demonstrates substantial improvements, with enhancements of 18.05 and 24.68 for flat-type and R-type images, respectively, which is a remarkable achievement.
\begin{table}[h]
\caption{ Comparisons of IOU on the thermal image dataset. $\left [ ^\dagger \right ]$ means the results are taken from~\cite{xue2023memory}. The best results are shown in bold.}
  \centering
  \begin{tabular}{lccc}
    \toprule
    Method& Flat-type & R-type \\
    \midrule
    3D-UNet++~\cite{hu2020lightweight} & 52.91 & 48.07 \\ 
    3D-UNet++-Sequence PCA~\cite{hu2020lightweight} & 66.38 & 61.77 \\
    ASGNet~\cite{Li2021AdaptivePL}$^\dagger$ & 52.36 & 46.40 \\
    \cite{xue2023memory}$^\dagger$ & 68.26 & 58.90 \\
    \hline
    SAM~\cite{kirillov2023segany} & 49.32 & 40.44 \\
    MedSAM~\cite{ma2023segment} & 66.56 & 65.75 \\ 
    DefectSAM (Ours) & \textbf{78.58} & \textbf{81.98}  \\
    \hline
    Improved ($\%$) & 18.05 & 24.68 \\ 
    \bottomrule
  \end{tabular}
\label{table-IOU}
\end{table}

\section{Conclusion}
This paper introduces DefectSAM, a novel approach for thermal image segmentation and defect detection. Leveraging the strengths of the pre-trained model SAM and a carefully curated thermal dataset, the proposed model attains SOTA performance when compared to existing image segmentation methods. DefectSAM demonstrates remarkable capabilities in detecting small and challenging defects within highly noisy thermal images, offering comprehensive and accurate results. Moreover, it exhibits favorable generalization ability, showcasing its potential to accelerate the advancement of defect detection tools.

\clearpage
{
    \small
    \bibliographystyle{ieeenat_fullname}
    \bibliography{main}
}
\clearpage
\setcounter{page}{1}
\maketitlesupplementary

\section{Experimental Specimen}
\label{specimen}
As depicted in Fig.~\ref{fig-system},  the heating process involves subjecting a specimen to heat, while the camera captures the corresponding temperature variations and transmits the data to the computer for analysis. The majority of the samples used in this study are carbon fiber reinforced polymers (CFRP), with a smaller number consisting of rubber and glass fiber materials. These samples exhibit internal defects that can be either artificially induced or naturally occurring. Furthermore, the samples exhibit diverse shapes, including planar types as well as irregular shapes such as curved and right angle bend (R-type) configurations. Variations in defect size and depth within the materials can also be observed. It is worth noting that during laboratory experiments, factors such as lamp power, sample positioning, and excitation duration contribute to the diversity of the acquired data. 
\begin{figure}[htbp!]
  \centering
   \includegraphics[width=0.9\columnwidth]{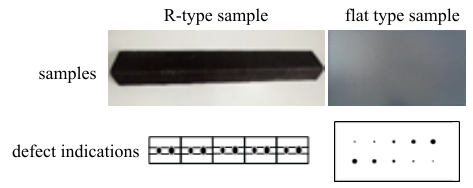}
   \caption{Two examples of samples exhibiting defects, each showcasing distinct variations in terms of shape, defect sizes, and depths.}
   \label{fig-sample}
\end{figure}

\section{Experimental Platform}
\label{platform}
\begin{figure*}[htbp!]
\centering
\includegraphics[width=0.9\textwidth]{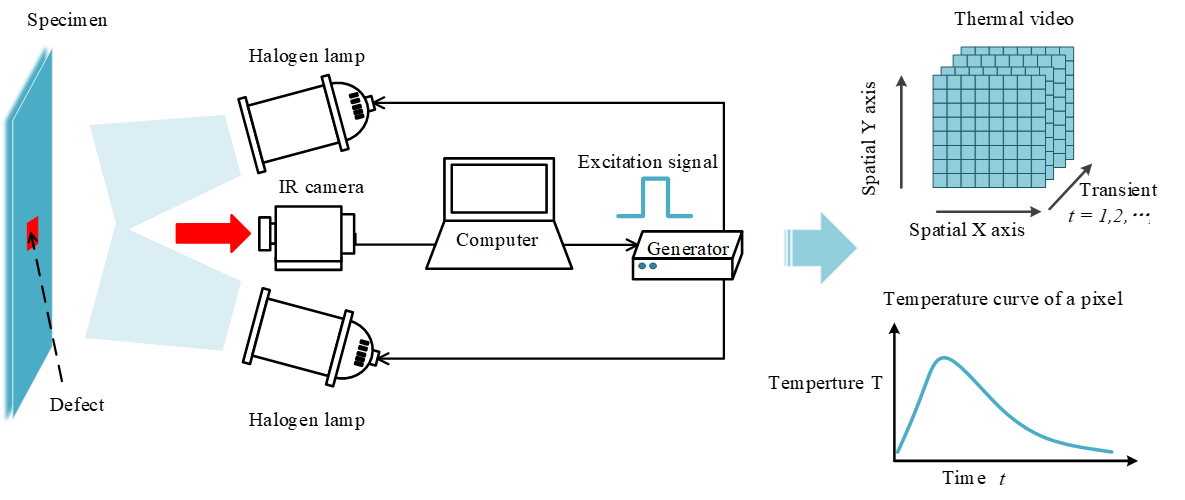} 
\caption{Diagram of the OPT system schematic, which illustrates the generation of thermal volumetric data for a specimen. At a specific point in this process, the temperature experiences a gradual increase followed by a subsequent decrease.}
\label{fig-system}
\end{figure*}

\begin{figure*}[htbp!]
\centering
\includegraphics[width=0.9\textwidth]{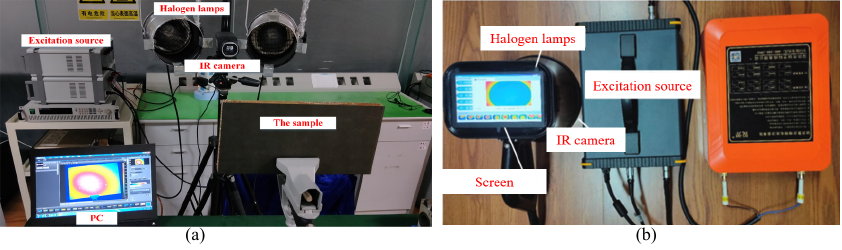} 
\caption{The laboratory experimental systems. (a) OPT system and (b) POPT system.}
\label{fig-OPT-POPT}
\end{figure*}

In this section, we present the Optical Pulsed Thermography (OPT) system and the Portable OPT (POPT) system. The schematic diagram of these systems is depicted in Fig.~\ref{fig-system}, and the corresponding physical diagrams for the OPT and POPT systems are shown in Fig.~\ref{fig-OPT-POPT}. These systems primarily comprise an excitation source, halogen lamps, a camera (FLIR A655sc), and a computer for processing and visualizing the generated three-dimensional thermal data. While the POPT system offers greater flexibility for outdoor use, it is limited by its lower resolution. The image in the bottom left corner of Fig.~\ref{fig-defect_types} is captured using the POPT system.

\end{document}